# Using compatible shape descriptor for lexicon reduction of printed Farsi subwords

Homa Davoudi
Dept. Electrical and Computer Engineering
Tarbiat Modares University
Tehran, Iran
h.davoudi@modares.ac.ir

Ehsanollah Kabir
Dept. Electrical and Computer Engineering
Tarbiat Modares University
Tehran, Iran
kabir@modares.ac.ir

*Abstract*— This Paper presents a method for lexicon reduction of Printed Farsi subwords based on their holistic shape features. Because of the large number of Persian subwords variously shaped from a simple letter to a complex combination of several connected characters, it is not easy to find a fixed shape descriptor suitable for all subwords. In this paper, we propose to select the descriptor according to the input shape characteristics. To do this, a neural network is trained to predict the appropriate descriptor of the input image. This network is implemented in the proposed lexicon reduction system to decide on the descriptor used for comparison of the query image with the lexicon entries. Evaluating the proposed method on a dataset of Persian subwords allows one to attest the effectiveness of the proposed idea of dealing differently with various query shapes.

## I. Introduction

Significant growth of textual Information in pictorial forms and the high cost of managing them create a great demand for the effective methods of automatic reading. Optical character recognition has been extensively studied in various applications and considerable progress has been reported. However, this field is still an active area of research, especially for processing of cursive scripts, like Persian and Arabic, where segmentation of the word image to its constitutive characters is more problematic. Although some segmentation methods are also proposed for this kind of scripts in previous studies, but the inherent complexity of this segmentation leads the researcher toward holistic methods, in which the initial segmentation are deleted.

In holistic approaches, features are directly extracted from the entire word image, considering it as a non separable entity. The main problem of this approach is the high number of classes, as the number of valid words to be recognized, i.e. lexicon size, may reach to 30000 words in unconstrained applications. So, in a lexicon-driven system where the expected transliteration of any input image is looked up in the lexicon, a large number of comparisons between the input image and all word hypotheses cause a high computational complexity. In addition, increasing the number of classes reduces the recognition accuracy. To deal with this problem, a lexicon reduction process is usually performed a-priori to reduce the number of word hypotheses. Performing a preliminary investigation, the lexicon reduction procedure tries to prune the lexicon by eliminating the unlikely subwords.

The lexicon reduction can be carried out either by using application-dependent knowledge (e.g. region names in postal applications) [1] or based on the characteristics of the query image itself (e.g. pruning visually dissimilar lexicon entries). The most informative characteristic frequently used in latter approach is word shape properties.

Mozaffari et. al. Ref. [2] proposed to use dots and diacritical marks in cursive handwritten Farsi/Arabic words for lexicon reduction. A technique was presented to extract the number and position of dots in a given word image, which are subsequently used to encode the word into strings. A string matching distance is then applied to compare the descriptor of query image with all lexicon entries and select the best matches. Another holistic method for handwritten Arabic lexicon recognition is proposed in [3], where the structural features of word shapes are encoded in a directed acyclic graph (DAG). Each graph is then mapped to a weighted topological signature vector space (W-TSV), where a nearest neighbors search is performed to achieve the reduced lexicon of any given image. Chherawala et. al. [4] introduced an sparse descriptor for lexicon reduction in handwritten Arabic documents which models the topological and geometrical features of subword images based on the skeleton pixel's local density. Adding up this information and diacritics count, their system generates an Arabic word descriptor to be used for indexing the reference lexicon, and consequently lexicon reduction. Ebrahimi [5] introduced a pictorial dictionary for eliminating search space in word recognition systems. In this work, which has been investigated on a set of 113,340 printed Farsi subwords, the pictorial dictionary is built by clustering the lexicon entries using the characteristic loci features extracted from each lexical word image. Given a query word image, this dictionary, accompanied with the minimum mean-distance classifier, is used to limit the considered words to the members of more probable clusters. Another method proposed in [6] for Farsi subwords uses the key character information along with the characteristic loci descriptor to organize the lexicon into the pictorial and textual dictionaries. These two organizations are combined to perform the lexicon reduction.

In this paper we present a new approach to reduce the lexicon size of printed Farsi subwords based on their holistic shape features, such that each subword is represented by a descriptor best suited to its shape characteristics. The idea comes from the fact that due to the large variety of subwords in an unconstrained lexicon, from a simple letter to the complex lengthy subwords (see Fig. 1), it is not practically feasible to find a fixed set of features (or generally a descriptor) being capable of properly capturing all these variations. So, we propose a system which, given an input image, first selects a descriptor structurally matched to its characteristics and then uses the selected descriptor in the reduction process.

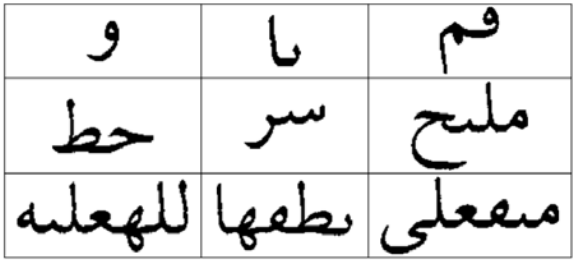

Fig. 1. Samples of subword images

Fig. 1 graphically depicts the main steps of our proposed method. Having selected a set of specific descriptors, during the training phase, the system learns to predict which descriptor is best compatible with a given subword image. This requires us to define what the best compatible descriptor is, and how to measure it. The compatibility of a descriptor with a subword image is defined as how well that descriptor can retrieve the subword image from a training set. According to this, the compatibility is quantified using a general retrieval evaluation measure. Computing the compatibility measures between a subword $s$ and every members of a predetermined set of descriptors $D=\{d_1,...,d_k\}$, the descriptor which gives the highest value is selected as the most compatible descriptor of $s$. This method is applied to all the training subwords, and the most compatible descriptor for each subword is specified and assigned to it. Using this in hand information, a neural network classifier is trained to predict the most compatible descriptors of any given image, among the set of descriptors. This network is implemented in the lexicon reduction system, where, once a query is presented, the system first specifies its most compatible descriptor and then, lexicon reduction is performed by making use of the selected descriptor.

In the following, we first describe in detail the proposed method of computing the compatibility measure in section 2. In section 3 we will describe the classifier designed to predict the proper descriptor of a query image. The overall reduction system will be presented afterward in section 4. The experiments are detailed in section 5 and Section 6 concludes the paper.

## II. SHAPE DESCRIPTOR

We employ three common global shape descriptors to construct D. Characteristic Loci [7], Fourier descriptor and zoning descriptor are the three descriptors we used here. These descriptors are implemented in [5] and it is claimed that the characteristic loci descriptor is the best among these three to be used for Farsi subwords. This assertion is demonstrated by comparing the correct classification rates obtained using each descriptor being applied on a test set of 5000 subwords. However, in this paper, instead of solely relying on one specific set of features, we propose to make use of all the descriptors such that, for every input image, the descriptor best suited to it will be used.

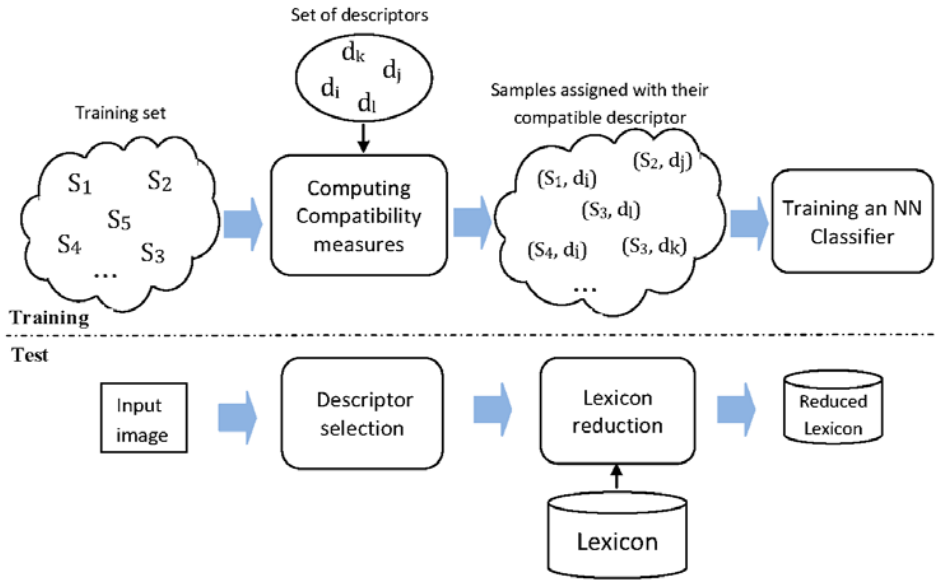

Fig. 2. Overall diagram of the proposed lexicon reduction system

These descriptors have been chosen according to their promising performance in describing Farsi subwords as reported previously. But it is noteworthy that as our proposed method is not directly dependent on the taken descriptors, this set can be replaced by any other set with sensible number of descriptors. However, in the experiments carried out here we are applying these three. In fact, our main focus here is to analysis the idea of learning the compatible descriptor for each image and its influence on the overall performance.

### III. COMPATIBILITY MEASURE

To quantitatively measure the compatibility of descriptor d with subword image S, a retrieval based procedure is applied. Accordingly, descriptor d is applied to retrieve the subword image S among all the training samples by sorting the training samples by their distances from S. The distance between two subword images is computed as a Euclidian distance between their feature vectors. The compatibility is then measured by evaluating the resultant retrieval list, measuring how well similar subwords as S appear near fronts of the list.

Among frequent evaluation metrics usually used to convert a retrieval list into a numeric score, we have selected discounted cumulative gain (DCG) [8] according to their stability shown previously. Unlike other methods limiting their scope of study into the first retrieved samples, DCG takes the whole retrieval list into account. To compute this measure, the matches within the same class as S are given a credit proportional to their distances from the top where matches near the front of the list will have more credits than ones near the end.

More formally, a binary list, $G$, is first formed to represent the correct matches of the retrieval list, where the elements $G_i$ is set to 1 if its corresponding element in list, $L_i$, is correctly retrieved and set to 0 otherwise. DCG measure is then computed recursively using (1).

$$DCG_i = \begin{cases} G_1 & i=1, \\ DCG_{i-1} + \dfrac{G_i}{\log_2(i)} & otherwise \end{cases} \quad (1)$$

The latest computed value ($DCG_N$) is finally normalized to the maximum possible value of DGC, and the DCG score is achieved as

$$DCG = \dfrac{DCG_N}{1 + \sum_{j=2}^{C} \dfrac{1}{\log_2(j)}} \quad (2)$$

Where C is the number of subwords similar to S, and N is the total size of the training set. So, DCG value gets a number between 0 and 1, such that a higher value implies better retrieval performance.

The DCG score of retrieving S from the training samples making use of descriptor d is considered as the compatibility measure, CM, of descriptor d with subword S.

Given a training sample S, the compatibility measure of every descriptor in D with respect to S is computed and the one with the highest value is assigned to S as its most compatible descriptor. In this way, each training subword is associated with one descriptor to be used for learning the classifier, as described in the following section

We perform this procedure for all subwords in our database. For each descriptors of D, subwords with highest CM values are presented in Fig. 3. This figure partly confirms our assumption that the compatible descriptor of shapes can be learnt. As can be implied from this figure, in each part, subwords share approximately similar features.

The CM values computed for various descriptors are compared in Fig. 4. Each curve in this figure corresponds to one descriptor, say $d_i$, which is drawn in descending order. (Notice that the horizontal axis in this figure does not convey any special data except the order of plotted data point which differs from a curve to another). According to this figures, all the descriptors are completely compatible with some subwords (i.e. every descriptors have some CM values equal to one.). However, the set of compatible subwords may differ from one descriptor to another (this cannot be inferred from this figure).

(a)

(b)

(c)

Fig. 3. Subwords given highest CM values for (a) Characteristic loci, (b) Fourier descriptor, (c) Zoning descriptor

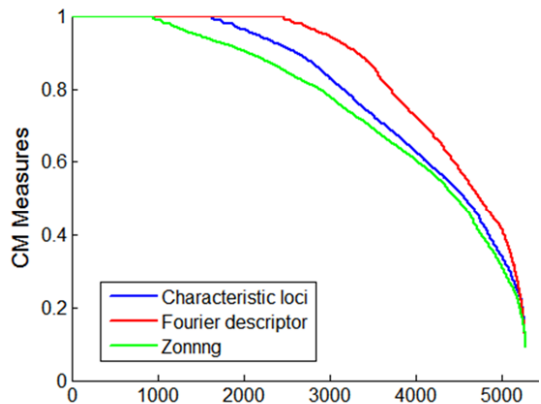

Fig. 4. Sorted CM values computed for each descriptors.

Another perceptible point about Fig.4 is the relatively higher CM values of Fourier descriptor which implies that, applying individually, Fourier descriptor provides the best retrieval result.

## IV. TRAINING THE CLASSIFIER

Selecting compatible descriptors for all training subwords, now the compatible descriptor for any input image should be predicted. As the input image is not known yet, it is not possible to use the retrieval based method presented in Sec. III to measure the CMs and select the descriptor with the highest value.

To address this issue, we use a neural network classifier to predict the most compatible descriptor for an input image. This solution relies on this reasonable assumption that each descriptor is good at capturing some specific shape characteristic, and selected as the compatible descriptor for the subwords with those characteristics. With this assumption, we expect to learn those specific characteristics using a neural network.

To this end, we implement a feedforward network, to classify the input image among 3 classes corresponding to the three descriptors. The set of features obtained by applying each descriptor are concatenated and applied as the input features of the network. The network is trained on the set of all training sample images. The trained classifier was tested and a correct classification rate of 71.5% was computed.

## V. LEXICON REDUCTION

After explaining the training phase, in this section, the entire proposed reduction system is presented. Given an input subword image and a lexicon, the main objective is to choose a number of subwords well-matched to the input subword as a reduce dictionary. To this end, features are extracted from the input image and compared with all lexicon subwords, to find the $n$ closest ones. These $n$ subwords are considered as the reduced lexicon. The value of $n$ is dependent on the required performance of the reduction in the intended application.

In addition to the relatively simple retrieval based method we use here for executing lexicon reduction, many more methods have been proposed in the literature that can be used as well. However, as our main focus here is on the effective shape description, we suffice to this simple method.

The novelty of our proposed reduction system is adopting the shape description according to the input image. This is done by initially presenting the input image to the trained neural network to select its suitable descriptor which is used to extract features from the image. These extracted features are compared with the same features of the lexicon entries to select the closer ones.

## VI. EXPERIMENTAL RESULTS

The database used in this paper consists of a set of 7000 images created by scanning a set of 1000 subwords printed in Lotus font with three different sizes of 10, 12 and 14. The printed subwords are then scanned in three resolutions of 40, 300 and 200 dpi. In this paper, ignoring the dots and diacritics, the features are extracted from the subword body shapes. Accordingly, the dots and diacritics are removed from the dataset images. Finally, the images are randomly divided into two sets of 5282 and 1718 images, respectively to be used as the training and test set. The training set is also served as the lexicon.

As the lexicon reduction method we used here is based on the image retrieval, the same measure as retrieval performance evaluation can be used here to assets the effectiveness of our proposed method. Accordingly, the DCG measures obtained by performing experiments to retrieve the samples of the test set from the lexicon using different descriptors are reported in Table I. Regarding the reported values, our proposed method of using suitable descriptor for any input image provides the best retrieval performance compared to each of individual descriptors.

For more detailed analysis, the precision-recall curves attained in performed experiments are presented in Fig. 5. The best performance of the proposed method is also clear in this figure. However, its superiority over the Fourier descriptor seems negligible, while it is not. The important fact here is the desired values of precision-recall which are dependent on the application. As our main goal is performing the lexicon reduction, the high recall rate is of much more importance. So, we need to focus on the corresponding regions of the curve. To make a better visualization, in Fig. 6 the recall rate is drawn versus another criteria called 'Degree of Reduction'[9], which is defined as the decrease in the size of the lexicon after reduction. This figure clearly shows the efficiency of our proposed method against others as, considering some specific degree of reduction, its recall values are significantly higher.

TABLE I.  COMPARING PERFORMANCE BASED ON THE DCG MEASURE VALUES

| Method | Characteristic Loci | Fourier descriptor | Zoning descriptor | Proposed Method |
|---|---|---|---|---|
| DCG Measure | 0.80 | 0.85 | 0.76 | 0.87 |

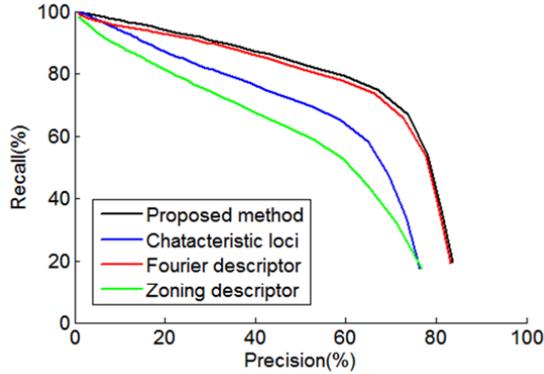

Fig. 5.  Performance evaluation: Recall vs. Precision

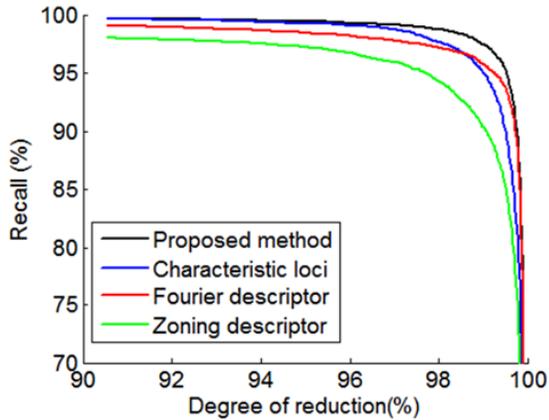

Fig. 6.  Performance evaluation : Recall vs. Degree of reduction

As the final experiment, we testify the effectiveness of the proposed method in comparison with the case that all descriptions features are concatenated and used simultaneously to retrieve the input image. This experiment is crucial to give an answer to this probable question: "The descriptor classifier of the proposed method uses all features extracted to decide on the compatible classifier. So, the effectiveness of this method may be due to the effect of employing various descriptors together". However, Fig. 7 explicitly declares that merely concatenating the features does not lead to the reasonable results.

## VII. CONCLUSION

In this paper, we presented a lexicon reduction method for printed Farsi subwords which, during the training phase, learns to predict the suitable descriptor for any given image. During the reduction, the system predicts the suitable descriptor of the input image and extracts the features accordingly. These features are then use for selecting the most likely lexicon entries. Applying the proposed method on a dataset of 7000 samples, its effectiveness was demonstrated.

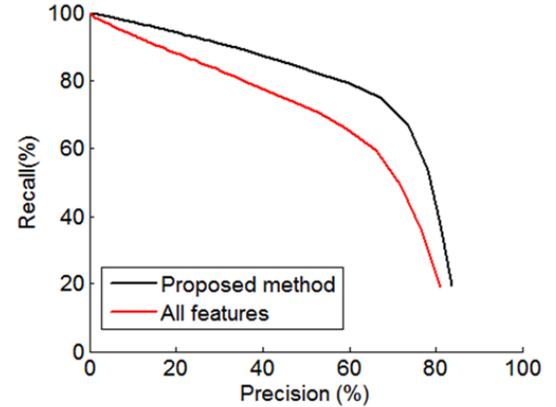

Fig. 7.  Comparing the performance of the proposed method with the the performance of reduction applying all features togather.